\documentclass[a4paper,fleqn, sn-standardnature,]{sn-jnl}
\jyear{2021}%
\usepackage{euscript}
\usepackage{subcaption}
\usepackage{mathtools}
\usepackage{amsmath}
\usepackage{url}
\usepackage[T1]{fontenc}

\begin{document}

\title [Solving drone flocking optimization using NSGA-II and PCA] {Drone Flocking Optimization using NSGA-II and Prinicipal Component Analysis}

\author*[1]{\fnm{Jagdish Chand} \sur{Bansal} \email{jcbansal@sau.ac.in}}
\author[1]{\fnm{Nikhil} \sur{Sethi} \email{sethi.nirvil@gmail.com}}
\author[2]{\fnm{Ogbonnaya}  \sur{Anicho} \email{anichoo@hope.ac.uk}}
\author[2]{\fnm{Atulya} \sur{Nagar} \email{atulya.nagar@hope.ac.uk}}

\affil[1]{ \orgname{South Asian University}, \orgaddress{\country{India}}}

\affil[2]{\orgname{Liverpool Hope University}, \orgaddress{\country{United Kingdom}}}

\abstract{
Individual agents in natural systems like flocks of birds or schools of fish display a remarkable ability to coordinate and communicate in local groups and execute a variety of tasks efficiently. Emulating such natural systems into drone swarms to solve problems in Defence, agriculture, industry automation and humanitarian relief is an emerging technology. However, flocking of aerial robots while maintaining multiple objectives, like collision avoidance, high speed etc. is still a challenge. In this paper, optimized flocking of drones in a confined environment with multiple conflicting objectives is proposed. The considered objectives are collision avoidance (with each other and the wall), speed, correlation, and communication (connected and disconnected agents). Principal Component Analysis (PCA) is applied for dimensionality reduction, and understanding the collective dynamics of the swarm. The control model is characterised by 12 parameters which are then optimized using a multi-objective solver (NSGA-II). The obtained results are reported and compared with that of the CMA-ES algorithm. The study is particularly useful as the proposed optimizer outputs a Pareto Front representing different types of swarms which can applied to different scenarios in the real world.
}


\maketitle

\section{Introduction}
% flocking literature
Collective behaviour is pervasive in nature and is frequently observed in diverse organisms ranging from microscopic bacteria \cite{allison_bacterial_1991} to large scale flocking of birds and insects\cite{nagy_hierarchical_2010}\cite{annurev-ento-010814-020627}. While researchers still hypothesize the underlying mechanisms behind such behaviour, moving in groups can offer several advantages like avoiding predators or carrying collective cargo\cite{ron2018bi}. Emulating these natural systems has gained popularity in the past few years and the development of a robust, fault-tolerant and generalised swarm of robots is now a widely regarded problem among researchers \cite{saffre2021design} \cite{coppola2020survey}. Aerial swarms, owing to their high maneuverability and speed, find a number of applications in various industries. They can be deployed as counterdrone measures \cite{brust_defending_2018} or basic surveillance operations in a defence scenario. In \cite{abraham2019swarm}, the authors utiltise the sensing capability of multiple robots to yield topographical and population density maps of a disaster afflicted area. In \cite{tosato_autonomous_2019}, a centralised swarm architecture was proposed for measuring air pollution. A system like this could potentially reduce measurement error due to the bigger sample size and distributed data points over the coverage volume. Mixed aerial and ground swarms have also been used for automation in construction tasks\cite{krizmancic_cooperative_2020}. In \cite{ju2018multiple}, multiple UAVs have been shown to outperform a single UAV for tasks like agricultural sensing and monitoring by measuring multiple metrics like energy consumption, flight time, and area coverage.
In general, drone swarms can be classified into three categories in order of increasing complexity: 
\begin{itemize}
    \item Coordinated: This refers to the collective movement with basic environmental awareness and collision avoidance.
    \item Cooperative: Here, the robots start to work together to achieve a certain goal using lesser resources than that of a single drone.
    \item Collaborative: This refers to multiple drones working together irrespective of their nature, i.e. heterogeneous collaboration.
\end{itemize}
In this paper, we propose a methodology to solve the drone swarm coordination problem with multiple conflicting objectives. In this document, the terms drone, UAV and aerial robot are used interchangeably.

% simulator literature 
Developing a robust velocity controller that allows multiple drones to self-organise comes with it's own set of challenges. According to the taxonomy defined in \cite{coppola2020survey} the control of velocities comes under Swarming Behaviour i.e. deciding a high level control policy with shared information across each agent's neighbors. The challenge is to take this shared information (the agent's own state as well states of neighbors) and come up with functions (policies) that output an instantaneous velocity vector for each agent. Over time, each agent's velocity gives rise to various patterns and mutual interactions that can potentially emerge into self-organising behaviour. Conventionally, a first of it's kind algorithm by Reynolds was based on simple rules for each agent and has been successfully applied in many fields \cite{hauert2011reynolds} \cite{dewi_simulating_2011} \cite{moere_time-varying_2004}. In \cite{vasarhelyi2018optimized}, the authors address this problem by defining a single fitness function and optimizing it through the CMA-ES algorithm. However, the study doesn't take into account multiple conflicting objectives, the priority of which can vary depending on the scenario. In \cite{marquez2021multi}, a multi-objective solution for quad-rotors is proposed but the swarm size is limited and the full range of solutions that take into account the relations among the fitnesses is not explored. We explore these relationships using unsupervised learning and extend our findings to highlight the use of obtaining a non-dominating solution set for drone flocking.

% simulations and literature study
Simulation is a powerful tool while studying these systems as it allows risk free experimentation and many programming platforms have been leveraged to come up with such solutions \cite{shah2018airsim}\cite{soria2020swarmlab}\cite{6920950}. In \cite{shah2018airsim}, the authors create a 3D simulator (AirSim\footnote{https://github.com/microsoft/AirSim}) for autonomous vehicles  written in C++ with growing support for multi-drone scenarios. It is easy to setup a simulation through Python scripting. The goal of AirSim however, is graphical realism and it is cumbersome to extend the same for the domain of swarm intelligence. RobotSim \footnote{https://github.com/csviragh/robotsim}, a fast simulator targeting flocking scenarios, is written in C but lacks sufficient abstraction and extension to multi-objective optimization which is required for the work carried out in this paper. In \cite{soria2020swarmlab}, the authors develop SwarmLAB\footnote{https://github.com/lis-epfl/swarmlab}, a dedicated drone swarm simulator capable of handling non-linear quadcopter simulation. While the simulator itself could meet requirements of speed and accuracy, it was insufficient for extending future work in the direction of hardware testing. We use Python to develop such a simulator as it can leverage existing libraries and algorithms. The inter-operability with libraries like Pymavlink, and the ease of testing protocols like the MultiWii Serial Protocol (MSP) on Raspberry pi on board computers was found to be particularly easy with Python based on past experience with similar projects.

% PCA literature study
Modelling of natural processes through simulation often needs to be complemented by an in depth qualitative understanding of the performance measures. Unsupervised learning can help in understanding and clustering data especially in high dimensional spaces which can't be visualised. It is widely used in experiments where abundant data is available such as mapping vulnerability indices \cite{abson2012using}, understanding relationships between economical and environmental objectives in a chemical supply chain \cite{pozo2012use}, understanding global motions of atoms in proteins \cite{loeffler2009collective}, and most commonly for dimensionality reduction in evolutionary algorithms \cite{deb2006searching}. Similar to many real systems, the solution of an optimization problem depends on various factors. Often, these factors or objectives are conflicting in nature and they cannot be solved simultaneously without compromising on the overall fitness. In case of flocking we consider six objectives: 
\begin{itemize}
    \item Collision avoidance with the wall.
    \item Collision avoidance with each other.
    \item Average speed of the swarm.
    \item Average velocity alignment or correlation.
    \item Total number of connected agents.
    \item Total number of disconnected agents.
\end{itemize}
We use PCA for both understanding collective dynamics of multi-agent systems and therefore reducing the objective space for the multi-objective optimizer. To the best of our knowledge, this work is the first attempt that involves using PCA to reduce the objective functions for a drone flocking optimization problem.
% Contrary to traditional objective reduction which discards all redundant objectives, we use a product of the non-conflicting fitnesses to make two non-redundant objectives. This allows us to compare the multi-objective fitness values easily with the single-objective fitness in \cite{vasarhelyi2018optimized}. 

These objectives are then optimized via a well-established multi-objective optimizer (NSGA-II) to yield a Pareto front which can be used to guide decision making and trade-offs under various situations. We report the results and show that the results at extremities of the Pareto front perform better than that of the CMA-ES algorithm. We conclude by giving some practical examples of the use of such abstract mathematical formalism for real time decision making with a flock of UAVs.

% Maint contributions and strucutre
In short, in this research, we create a drone swarm simulator integrated with a multi-objective solver, use PCA to understand the collective dynamics of swarms, and give a Pareto front that represents different swarms that can be used in real-world scenarios. 
The rest of the paper is organised as follows: Section 2 presents the background of Principal Component Analysis (PCA) and multi-objective optimizer (NSGA-II). A Drone flocking optimization problem is formulated in Section \ref{simulation_framework}. In Section \ref{pca} PCA is used to reduce the number of the objective functions and a discussion on the correlations is followed. Section \ref{setup} presents the experimental setup while the numerical results and discussions are given in Section \ref{results}. The research is concluded by giving some potential use-cases and possible future work.   

\section{Background} \label{background}
\subsection{Principal Component Analysis} \label{background_pca}
A high dimensional objective space is known to suffer from problems like poor selection pressure and convergence\cite{deb2006searching}. It is also difficult to visualize the space and gain intuition which is often required for appropriate decision making. 
Principal component analysis, a technique under the domain of unsupervised learning may be useful to understand the underlying structure of the data without explicit labels. The idea is to search for the eigenvectors of an $m$-dimensional covariance matrix ($K$) which is then used to decide the redundant objectives. Here, $m$ is the number of objectives. This covariance matrix (often called correlation matrix when scaled) is symmetric and it's elements give the relations between the design variables on which the analysis has been run. Such an analysis on the objectives of a optimization problem can give insights about their correlations and can help in understanding their qualitative aspects.

%  Such information is often used to make important global decisions with respect to the scenario that the UAVs face. 
Let $X$ be an $n$  x  $m$ design matrix with $n$ rows as the samples and $m$ columns as objectives. A pre-processing step often carried out is the normalisation of design matrix to bring the mean of samples for each objective to 0 and the variance to 1.0 (Eq. \eqref{eq_featurescaling}). The covariance matrix is then calculated by taking the mean of all samples of the pairwise products for each objective (Eq. \eqref{eq_cov_element}). In a vectorized format, this is equivalent to taking the matrix product of the design matrix $X$ with it's transpose (Eq. \eqref{eq_covarmat}).
\begin{align}
    %feature scale
    X_{ij}^{norm} &= \frac{X_{ij}-\mu_{j}}{\sigma_{j}} \label{eq_featurescaling}\\
    %covariance element
    K_{ij} &= \frac{1}{n} \sum^{n}_{k=1} X_{ki}X_{kj} \label{eq_cov_element}\\ 
    %covariance matrix
    K &= \frac{1}{n} (X^{norm})^{T} X^{norm} \label{eq_covarmat}
\end{align}
\begin{flushleft}
    where,\\
    $X_{ij} \ \equiv$ Element of $X$ at $i^{th}$ row and $j^{th}$ column \\
    $X_{ij}^{norm} \ \equiv X_{ij}$ normalised to 0.0 mean and 1.00 standard deviation \\
    $\mu_{j} \ \equiv$ Mean of all $n$ samples of $j^{th}$ objective \\
    $\sigma_{j} \ \equiv$ Standard deviation of all $n$ samples of $j^{th}$ objective \\
    $n \ \equiv$ Number of samples \\
    $m \ \equiv$ Number of objectives \\
    $K \equiv $ Covariance matrix\\
    $K_{ij} \ \equiv$ Element of $K$ at $i^{th}$ row and $j^{th}$ column \\
\end{flushleft}

\subsection{Non-Dominating Sorting Genetic Algorithm-II} \label{background_nsga}
NSGA-II is a multi-objective optimization algorithm that is based on ranking each solution in the population according to their fitness and progressively producing better solutions using genetic operators like reproduction and mutation. The entire algorithm is explained in detail in \cite{deb_fast_2002}. However, a brief explanation covering the salient features of NSGA-II is explained here.

Let $P_{o}$ be a N sized initial random  population. This population is sorted based on non-domination according to the following rules:
An individual $X_{1}$ in the population is said to be dominated by individual $X_{2}$ if satisfies both of the following conditions:
\begin{itemize}
    \item  All fitnesses of $X_{1}$ must be less than or equal to that of $X_{2}$ particle.     
    \item At  least one fitness of $X_{1}$ must be strictly less than that of $X_{2 }$.
\end{itemize}

Mathematically, individual $X_{1}$ dominates $X_{2}$ if $d = 1$, and the individuals are non-dominated if $d=0$.
\begin{center}
    Where, $d = \{ \forall m \ F(X_{1})^m \leq F(X_{2})^m \} \cap  \{ \exists m \ F(X_{1})^m < F(X_{2})^m \}$    
\end{center}

This method  divides the population into dominating and non-dominating solutions which is a heuristic used to guide the population towards better solutions through the generations. Each solution in this population is also ranked based on the number of other members it is dominated by and accordingly it is assigned a front rank. 
Next, an offspring population Q is created from the sorted population by applying tournament selection, recombination and mutation operators. To ensure that the best solution across generations remains (elitism), a new 2N sized population is made using $P \cup Q$ which is again sorted and ranked.  To make the next population $P_{t+1}$ from this combined set, solutions are taken in order of their front ranking. In case the number of solutions belonging to a front exceed the amount that can be accommodated into the new N sized  population, the remaining solutions in that front are ranked based on a crowding operator as follows:

Let $\mathcal{F}^{k}$ be the set of solutions on the $k^{th}$ ranked pareto front. The crowding distance ($c^{m}_{i}$) for the $m^{th}$ objective for $i^{th}$ solution on this front is defined as the normalised distance between the two nearest solutions i.e. $(i+1)^{th}$ and $(i-1)^{th}$ (Eq. \eqref{eq_crowd_im}). The overall crowding distance ($c_{i}$) is the sum taken for each objective (Eq. \eqref{eq_crowd_i}).
\begin{align}
    \forall X_{i} \in \mathcal{F}^{k}: c^{m}_{i} &= \frac{F^{m}(X_{i+1})-F^{m}(X_{i-1})}{F^{m}_{max}-F^{m}_{min}} \label{eq_crowd_im}\\
    c_{i} &= \sum^{M}_{m=1} c^{m}_{i} \label{eq_crowd_i}
\end{align}

This crowding operator ensures that the Pareto Front is uniformly distributed and the range of each each objective value is maximised as the search progresses. The remaining solutions are ranked according to $c_{i}$ and the new population $P_{t+1}$ moves forward to the next generation.
NSGA-II is faster than NSGA-I and has a worst case complexity of $O(MN^{2})$. 
\section{Drone flocking optimization problem} \label{simulation_framework}
A completely decentralized flocking swarm is based on certain simple rules like Separation, Alignment, Cohesion. These rules when defined using a velocity control algorithm has certain parameters which can be tuned to flock optimally. In this section these parameters are introduced and a  simulation framework capable of handling artificial sensor noise is created. The algorithm used for flocking is based on the work in \cite{vasarhelyi2018optimized} and Reynold's Flocking model \cite{10.1145/37402.37406}. Some subtle modifications have been incorporated in order to handle a multi-objective optimization framework. We use vectorized versions of the equations to leverage fast computation with matrix computation libraries.

To simulate a multi-agent system, there must be a mechanism to share information across the agents. In case of a decentralised system this information is shared in each agent's neighborhood $\EuScript{N}_{o}$. Moreover, real-systems are characterised by a stochastic uncertainty and noise which are incorporated into the position($\textbf{r}$) and velocity($\textbf{v}$) vectors of the drones. The model for simulating the noise is taken from \cite{viragh2014flocking}. The relative position ($\textbf{r}_{ji}$) and velocity ($\textbf{v}_{ji}$) at time $t$ is then found using the following equations:

\begin{align}
    \textbf{r}_{ji}(t) &=(\textbf{r}_{j}(t-t_{del})+\textbf{r}_{j}^{gps})-\textbf{r}_{i}(t)-\textbf{r}_{i}^{gps}\\
    \textbf{v}_{ji}(t) &=(\textbf{v}_{j}(t-t_{del})+\textbf{v}_{j}^{gps})-\textbf{v}_{i}(t)-\textbf{v}_{i}^{gps} \\
    R^{rel}_{j} &= \textbf{r}_{ji}(t)\\
    V^{rel}_{j} &= \textbf{v}_{ji}(t)
\end{align}

\begin{flushleft}
    where,\\
    $\textbf{r}_{ji}\ \equiv$ Relative position vector of $j^{th}$ agent with respect to $i^{th}$ agent at time $t$\\
    $\textbf{v}_{ji}\ \equiv$ Relative velocity vector of $j^{th}$ agent with respect to $i^{th}$ agent at time $t$\\
    $R^{rel}_{j} \ \equiv$ $j^{th}$ row of the Relative position matrix for agent $i \ \forall \ j = {1,2,...,\EuScript{N}_{o}}$\\
    $V^{rel}_{j} \ \equiv$ $j^{th}$ row of the Relative velocity matrix for agent $i \ \forall \ j = {1,2,...,\EuScript{N}_{o}}$\\
    $t_{del} \ \equiv$ Simulated communication delay \\
    $\textbf{r}^{gps} \ \equiv$ Simulated GPS noise for position\\
    $\textbf{v}^{gps} \ \equiv$ Simulated GPS noise for velocity
\end{flushleft}

\subsection{Decision variables} \label{sec:variables}
Based on the above modification for the relative position and velocities the flocking rules are explained the following sections. These rules give rise to certain parameters which are used as decision variables for the drone flocking optimization problem. Note that all the flocking operations are carried out for all $N$ agents .

\subsubsection{Separation} \label{sec:seperation}
To flock effectively without collisions, the agents must have a mechanism for repulsion. A spring-like mechanism is used which is activated at short ranges of inter-agent distance in the flock. The following two equations \eqref{eq_Vrep} and \eqref{eq_vrep} are used to find a repulsion vector for agent $i$ after scaling it according to the relative distances in $\textbf{r}_{i}^{mag}$ and a gain $p^{rep}$.

\begin{align}
    \textbf{r}^{mag} &= \|R^{rel}\| ^{\bot r_{0}^{rep} } \label{eq_rmag}\\
    V^{rep} &= p^{rep}. (\textbf{r}^{mag} - r_{0}^{rep}).\frac{R^{rel}}{\textbf{r}^{mag}} \label{eq_Vrep}\\
    \textbf{v}^{rep}_{i} &=\sum_{j=1}^{\EuScript{N}_{o}} V^{rep}_{j} \label{eq_vrep}
\end{align}

\begin{flushleft}
    where,\\
    $\ a^{\bot c}_{\top b} = min(max(a,b),c))$\\
    $\textbf{r}^{mag} \equiv \EuScript{N}_{o} \ \text{x} \ 1 $  sized vector containing inter-agent distances \\
    $r_{0}^{rep} \ \equiv$ Repulsion cutoff distance (user dependent parameter)\\
    $p^{rep} \ \equiv$ Repulsion gain (user dependent parameter)\\
    $V^{rep} \ \equiv  \EuScript{N}_{o} \ \text{x} \ 2 $ sized matrix of scaled repulsion velocities \\
    $V^{rep}_{j} \ \equiv$ Repulsion velocity of $j^{th}$ neighbor \\
    $\textbf{v}^{rep}_{i}\ \equiv$ Desired collective repulsion vector 
\end{flushleft}

Note that the upper bound of $\textbf{r}^{mag}$ is the parameter $r_{0}^{rep}$ to enable short range effects. The matrix norm in Eq. \eqref{eq_rmag} is only taken along the row axis, i.e. for each neighbor. $V^{rep}$ contains all the the corresponding scaled repulsion velocities and the division and multiplication in Eq. \eqref{eq_Vrep} is done element wise. 
% A summation along the column axis in in equation \ref{eq_vrep} results in the repulsion velocity $\textbf{v}_{i}^{rep}$ for the concerned drone. The same notation and theme is followed in the following subsections.

\subsubsection{Alignment} \label{sec:alignment}
V{\'a}s{\'a}rhelyi et al. \cite{vasarhelyi2018optimized} realised that effective control of \emph{both} the magnitude and direction of velocities as a function of inter-agent distances can yield the best alignment with scalable velocities. The equations for alignment are similar to repulsion with one major difference: the upper bound for the velocity magnitude ($\textbf{v}^{frictmax}_{i}$) is now calculated dynamically with decay function $D$ in Eq. \eqref{eq_vfrictmax} which is dependent on the inter-agent distance \cite{vasarhelyi2018optimized}. Eqs. \eqref{eq_vmag} - \eqref{eq_vfrict} describe the process of finding out the combined alignment vector for the agent.

\begin{align}
    \textbf{v}^{frictmax} &= D(\textbf{r}_{i}^{mag}- r_{0}^{frict} - r_{0}^{rep}, a^{frict}, p^{frict}) _{\top {v}^{frict}} \label{eq_vfrictmax}\\
    \textbf{v}^{mag} &= \|V^{rel}\| _{\top \textbf{v}^{frictmax}} \label{eq_vmag} \\
     V^{frict} &= c^{frict}.(\textbf{v}^{mag}-\textbf{v}^{frictmax}).\frac{V^{rel}}{\textbf{v}_{mag}} \label{eq_Vfrict}\\
    \textbf{v}^{frict}_{i} &=\sum_{j=1}^{\EuScript{N}_{o}} V^{frict}_{j} \label{eq_vfrict}
\end{align}
\begin{flushleft}
    where,\\
    $D$ is a vectorized version of the velocity decay function taken from \cite{vasarhelyi2018optimized}\\
    $p^{frict} \equiv$ Slope for the linear part of the decay curve (user dependent parameter)   \\
    $a^{frict} \equiv$ Acceleration for the non-linear part of the decay curve (user dependent parameter)   \\
    $c^{frict} \equiv$ Overall Gain for alignment (user dependent parameter)   \\
    $v^{frict} \equiv$ Velocity slack for alignment (user dependent parameter) \\
    $r_{0}^{frict} \ \equiv$ Alignment cutoff distance for maximum alignment (user dependent parameter)\\
    $\textbf{v}^{frict}_{i}\ \equiv$ Desired collective alignment vector
\end{flushleft}

Eq. \eqref{eq_vfrictmax} gives a vector composed of the maximum allowable velocity difference for each neighbor. The maximum is proportional to the inter-agent distance. This ensures that the alignment for two agents that are in close proximity is larger and vice-versa. Also, the maximum allowable difference is lower bound by an optimization parameter $v^{frict}$ so that the agents do not strive for perfect alignment and there is some slack.
Eq. \eqref{eq_Vfrict} compensates the velocity difference for each neighbor and Eq. \eqref{eq_vfrict} sums the alignment velocities for each neighbor.

\subsubsection{Wall collisions} \label{sec:collisions}
To account for collisions at walls, the authors in  \cite{vasarhelyi2018optimized}  have proposed virtual "shill" agents at the walls which the actual agents can try to align their velocities with.
These shill agents have no gain and therefore repulsion at walls takes place to the maximum extent ($c^{shill}=1$). This makes sense while flocking in confined environments, because one of the primary goals is to avoid the wall at any cost. In our research however, while seeking for non dominated set of solutions(ref. section \ref{results}), we can characterise the \emph{elasticity}  of the virtual geo-fence using a shill gain ($c^{shill}$) parameter. 
The following equations are used to find a shill velocity vector from each wall so as to align with it. $\textbf{r}_{ci}$ is the relative position vector from the agent to the arena's center $\textbf{r}_{c}$. This vector is used to find the distances to the walls in Eq. \eqref{eq_rsmag}. Eq. \eqref{eq_Vs} gives a $m \ \text{x} \ m$ sized matrix which has the rows as the shill vector from each wall. 

\begin{align}
    \textbf{r}_{ci} &= \textbf{r}_{c} -\textbf{r}_{i} \label{eq_rci} \\
    \textbf{r}_{s}^{mag} &= L_{c}/2 - \vert \textbf{r}_{ci} \vert \label{eq_rsmag}\\
    \textbf{v}^{shillmax}_{i} &= D(\textbf{r}^{mag}_{s}- r_{0}^{shill}, a^{shill}, p^{shill}) \\
    V_{s} &= (v^{shill} . \frac{\textbf{r}_{ci}}{\vert \textbf{r}_{ci} \vert}) \odot I \label{eq_Vs}\\
    \textbf{v}_{s}^{mag} &= \| V_{s} - \textbf{v}_{i} \| \ _{\top \textbf{v}^{shillmax}_{i}} \label{eq_vsmag} \\
    V^{shill} &= c^{shill}.(\textbf{v}_{s}^{mag}-\textbf{v}^{shillmax}_{i}).\frac{V_{s}}{\textbf{v}_{s}^{mag}} \\
    \textbf{v}^{shill}_{i} &= \sum_{k=1}^{m} V^{shill}_{k} \label{eq_vssum}
\end{align}
\begin{flushleft}
    where,\\
    $\textbf{r}_{c}$ Absolute position of the center of the arena\\
    $L_{c}$ Side length of the arena\\
    $\textbf{r}_{ci}$ Relative position of the center with respect to the agent\\
    $p^{shill} \equiv$ Slope for the linear part of the decay curve (user dependent parameter)   \\
    $a^{shill} \equiv$ Acceleration for the non-linear part of the decay curve (user dependent parameter)   \\
    $c^{shill} \equiv$ Overall Gain for shilling alignment (user dependent parameter)   \\
    $v^{shill} \equiv$ Speed of shilling agents (user dependent parameter) \\
    $r_{0}^{shill} \ \equiv$ Alignment cutoff distance for maximum alignment (user dependent parameter)\\
    $\textbf{v}^{frict}_{i}\ \equiv$ Desired collective alignment vector
\end{flushleft}
We assume a square geo-fence in our research but trivial modifications to Eq. \eqref{eq_rci} and \eqref{eq_rsmag} can generalise it other shapes as well. Here, $I$ is the identity matrix and $\odot$ is the Hadamard product. Eqs. \eqref{eq_vsmag} - \eqref{eq_vssum} have the same velocity alignment procedure done in section \ref{sec:alignment} but here it's done for each wall's shill velocity instead of each agent. 

The above three velocities (\ref{sec:seperation} - \ref{sec:collisions}) along with the normalised flocking velocity are summed up and normalised again to give the desired velocity for the respective agent.
\begin{align}
    \textbf{v}_{i}^{desired} &= \frac{\textbf{v}_{i}}{\| \textbf{v}_{i} \|} v^{flock} + \textbf{v}_{i}^{rep} + \textbf{v}_{i}^{frict} + \textbf{v}_{i}^{shill}\\
    \textbf{v}_{i}^{desired} &\longleftarrow min\{v^{max}, \| \textbf{v}_{i}^{desired} \| \} \ \frac{\textbf{v}_{i}^{desired}}{\| \textbf{v}_{i}^{desired} \|}
\end{align}

Finally, the set of resulting 12 parameters to optimize is:
\begin{equation*}
    x = \{ r_{0}^{sep}, p^{rep}, r_{0}^{frict}, a^{frict}, p^{frict}, v^{frict}, c^{frict}, r_{0}^{shill}, v^{shill}, a^{shill}, p^{shill}, c^{shill} \} 
\end{equation*}

\subsection{Fitness functions} \label{sec:fitfuncs}
To measure the performance of one simulation run, order parameters are defined and passed through transfer functions to get the fitnesses\cite{vasarhelyi2018optimized}.

\begin{align}
    \label{eq_fitfuncs}
    \begin{split}
    F^{speed} &= \textbf{F}_{1}(\phi^{vel}, v^{flock}, v^{tol}) \\
    F^{coll} &= \textbf{F}_{3}(\phi^{coll}, a^{tol}) \\
    F^{wall} &= \textbf{F}_{2}(\phi^{wall}, r^{tol}) \\
    F^{corr} &= \Theta(\phi^{corr})\phi^{corr} \\
    F^{disc} &= \textbf{F}_{3}(\phi^{disc}, N/5) \\
    F^{cluster} &= \textbf{F}_{3}(\phi^{cluster}, N/5)     
    \end{split}
\end{align}  

\sloppy
% \hfill
Here, the order parameters $\phi^{vel}, \phi^{coll}, \phi^{corr}, \phi^{wall}$ and transfer functions  $\textbf{F}_{1}$, $\textbf{F}_{2}$, $\textbf{F}_{3}$ are taken from \cite{vasarhelyi2018optimized} and $\Theta$ is a the heave-side step function. Parameters $r^{tol}, a^{tol}, \text{and } v^{tol}$ are explained in section \ref{setup}. Order parameters for disconnected agents ($\phi^{disc}$) and the minimum connected agents ($\phi^{cluster}$) are explained below. These parameters are calculated locally in $r^{cluster}$ sized clusters:
\begin{align}
    r^{cluster} = r^{rep} + r^{frict} + \tilde{D}(v^{flock}, a^{frict}, p^{frict})
\end{align}
$\tilde{D}$ is the braking distance r for which D(r, a, p) = v for any agent.

\subsubsection*{Disconnected agents}

This parameter measures the average number of completely disconnected agents  throughout the simulation. Eq. \eqref{eq_ncluster} gives the number of agents within $r^{cluster}$ distance of each agent at any given moment. Eq. \eqref{eq_phidisc} is then used to determine the number of agents throughout the simulation with zero connected agents, i.e. disconnected.

\begin{align}
    \Theta(x) &= \begin{cases}
                    1 & \text{if $x \geq 0$} \\
                    0 & \text{if $x < 0$}
                 \end{cases} \\
    \\
    n_{i}^{cluster}(t) &= \sum_{j \neq i}^{N-1} \Theta(r^{cluster} - r_{ij}(t)) \label{eq_ncluster}\\
    \phi^{disc} &= \frac{1}{T} \int_{0}^{T} \sum_{i=1}^{N} \Theta (n_{i}^{cluster}(t)-1) \label{eq_phidisc}
\end{align}

\subsubsection*{Minimum connected agents}
This parameter measures the minimum number of connected agents averaged throughout the simulation and is therefore dependent on time. Since the drones start at random positions, it was observed that keeping this parameter time dependent instead of steady state (the global minimum throughout the simulation) gave a better idea of the robustness of the communication graph throughout the simulation.

\begin{align}
    \phi^{cluster}(t) &= \frac{1}{T}\int_{0}^{T} min \{n_{1}^{cluster}, n_{2}^{cluster}... \ \ n_{i}^{cluster} \} (t)
\end{align}

\begin{flushright}
    $\forall \ i: 1,2,...,N$
\end{flushright}
\hfill\\
Finally, we need to optimize these fitness functions given in Eq. \eqref{eq_fitfuncs} simultaneously
To optimize the above fitness functions simultaneously, the six objectives must be analyzed for correlations among them so that the system can be represented with fewer objectives, preferably two. In the next section, Principal Component Analysis (PCA) is used for the dimensionality reduction so that the multi-objective optimizer NSGA-II can be used, effectively.

\section{Dimensionality reduction using PCA} \label{pca}
To reduce the number of objective functions, a data set of the six objectives discussed in section \ref{simulation_framework} is collected. This data is just the result of 500 random simulations without any heuristic so as to cover the entire search space. Note that the data used for PCA is for the fitness values after being passed through the transfer functions. This can also be done directly on the order parameters as well. Both processes would give different correlations depending on the nature of the transfer function. We prefer the former method as it gives a more accurate representation of the matrix components and the exact fitnesses functions used for optimization. 
This data is used to create the covariance matrix and principal components shown in section \ref{background} which is followed by a qualitative discussion on the correlations.

\begin{figure*}
  \begin{minipage}{.1\linewidth}
    \centering
    \phantom{M} \\
    \phantom{M} 
    \[\begin{array}{c}
        F^{wall}\\
        F^{speed}\\
        F^{corr}\\
        F^{coll}\\
        F^{disc}\\
        F^{cluster}
    \end{array}\]
  \end{minipage}
  \begin{minipage}{.2\linewidth}
    \centering
    $\ \ 1^{st}$ component \\ $(w)$
    \[\left[\begin{array}{c}
      0.191\\
    0.329\\
    -0.495\\
    -0.285\\
    -0.509\\
    -0.518
    \end{array}\right]\]
  \end{minipage}
  \begin{minipage}{.7\linewidth}
    \centering
    Covariance matrix \\$(K)$
    \[\left[\begin{array}{cccccc}
      1.002&0.2821&-0.172&-0.115&-0.094&-0.131\\
    0.282&1.002&-0.285&-0.283&-0.301&-0.259\\
    -0.172&-0.285&1.002&0.253&0.5815&0.682\\
    -0.115&-0.283&0.253&1.002&0.278&0.204\\
    -0.094&-0.301&0.581&0.278&1.002&0.748\\
    -0.131&-0.259&0.682&0.204&0.748&1.002
    \end{array}\right]\]
  \end{minipage}
    \caption{Matrices obtained from Principal component analysis}
    \label{fig_mat_pca}
\end{figure*}

In Fig. \ref{fig_mat_pca}, the matrices obtained from the application of PCA on the objective space is given. The matrices show some interesting results. Some insights are discussed as follows:

$K_{23}$ is negative implying that a higher velocity doesn’t necessarily imply higher correlation. This  might be false in situations where the UAVs have very high velocity magnitudes while travelling long distances or have a large turn radius (as in the case of fixed wing drones). But in a confined environment, to maintain correlation at the edges (where the flock gets broken up most), the speeds must be reduced. This is also a consequence of a limited acceleration which aligns with the actual physical systems.

The above statement regarding confined environments is also confirmed by $K_{13}$. To maintain correlation at walls, the UAVs can either slow down or skip the wall altogether. A combination of slowing down and breaching the geo-fence makes the above movement the most efficient. Note that intuition would suggest that as speed increases, it would be easier to decrease the wall fitness as there is indeed a limited acceleration/deceleration available. Upon running numerous simulations and making correlation matrices, it was found that this is because $F_{wall}$ itself is time dependent. This means that the fitness is inversely proportional to the amount of time frames that the drones spend outside the wall. Since the goal of $F_{2}$ is to maintain correlation and connectivity wherever possible at the expense of $\phi^{vel}$ and $F^{wall}$, whenever the drones slow down they naturally spend more time frames outside and turn slowly irrespective of the acceleration, This makes $F^{wall}$ and $F^{speed}$ directly correlated with each other on average. The elements $w_{11}$ and $w_{12}$ and $K_{12}$ are representative of this very fact.

Drones naturally collide less with each other when their velocities are aligned and they are well connected. This is because the time it takes for velocity changes to travel throughout the communication network is much lesser. Although, when this network is strongly connected, the agent has to sum up through many velocity differences in it's neighborhood. While this is advantageous when the neighbors are moving in similar directions, it can be detrimental when there is a lot of noise and the inter-agent velocity differences point in different directions. As a result the summed up alignment velocity for the concerned agent gets dampened by cancelling out. This results in inter-agent friction and makes the entire flock sluggish (slow to react). This is very clearly shown by the elements $w_{2}$ and $w_{3}$ which are strongly uncorrelated. It is also worth pointing out that elements $w_{3}$ through $w_{6}$ are strongly correlated which confirms the association of correlation and collisions with the communication network. The cluster parameters for disconnection and minimum number of connected UAVs are strongly correlated ($K_{65}$) as expected as they are both direct functions of the communication network.

Using a single objective can result in loss of important information as the final fitness is just the collective product or weighted sum. Particularly, in noisy dynamical systems such as multi-agent robotics, efforts need to be made to retain as much information as possible and use it intelligently to guide the decision making process. We propose a multi-objective methodology for optimization of the swarm’s fitness to tackle this problem.

The principal component ($w$) for the maximum variance captures all the above relations and shows them how they relate with each other on average. The sign of the elements indicate correlation which gives rise to the following features/objectives:

\begin{align}
    F_{1} &= F^{wall} \ . \ F^{speed}\\
    F_{2} &= F^{corr} \ . \ F^{coll} \ . \ F^{disc} \ . \ F^{cluster}
\end{align}

Unlike traditional PCA, we don't use just the non-redundant objectives. Each objective captures tangible physical information about the simulation and therefore we multiply the two sets individually to retain that information and also make it easier to draw a comparison with the single objective CMA-ES optimizer as given in section \ref{results}.

It is worth mentioning here that the above correlation matrix is dependent on the number of agents and the size of the confined arena. While some parameters like the cluster connectivity and correlation still remain the same because they are independent of the above parameters, a different non redundant set of objectives was obtained upon changing the size of the geo-fence. The simulations dictated  that the same number of agents in a larger space took more time to align with the shill agents due to the stronger inter-agent alignment over long distances. The correlation matrix for the same is not shown here for the sake of brevity. In \cite{vasarhelyi2018optimized}, there is a certain ambiguity in the size of the geo-fence. While the authors mentioned that they used a side length of 250m for the square arena, the averaged results on their open-source simulator were closer to the claimed ones when a radius of 250m (or side length of 500m) was used for the arena. To make comparisons easier we continue with the latter definition for our study as well.

The reduced objectives are passed to the multi-objective solver NSGA-II \cite{deb_fast_2002} and the results are summarized below.

\section{Numerical Experiments } \label{setup}
\begin{figure}
    \centering
	\includegraphics[scale=0.35]{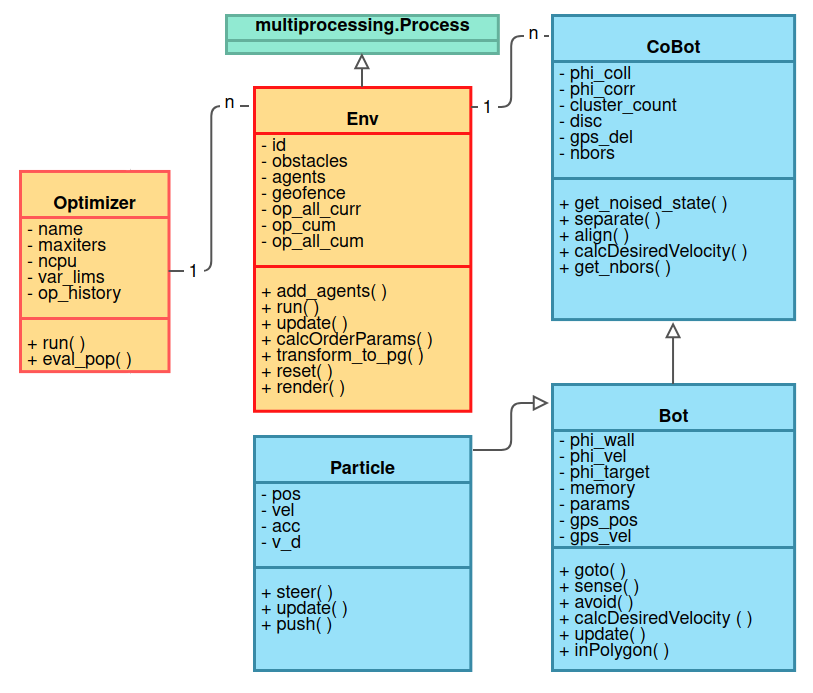}
	\caption{Class diagram}
	\label{fig:class_diagram}       % Give a unique label
\end{figure}

\begin{figure}
    \centering
	\includegraphics[scale=0.3]{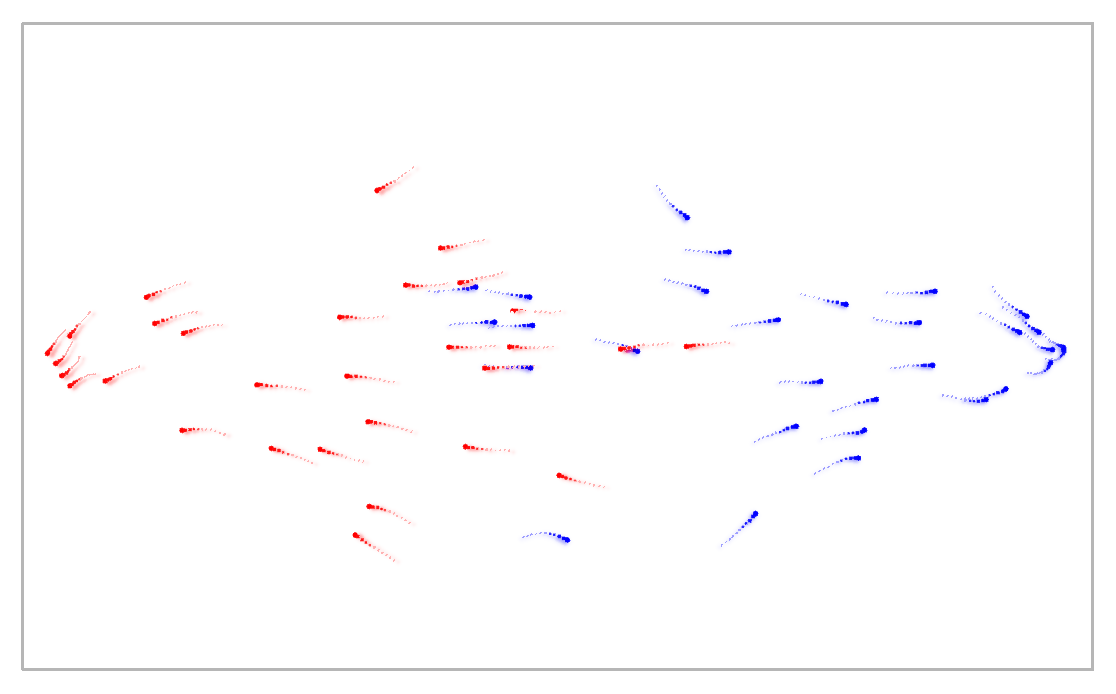}
	\caption{MOflock simulation screenshot}
	\label{sim_screenshot}       % Give a unique label
\end{figure}

To test the propsed algorithm and for future work as well, a custom simulator MOflock was created in the Python programming language. The ease of use in setting up multiple processes, leveraging optimization and machine learning libraries was a major influence in choosing Python. The simulator is highly object oriented and modular. It has the drones abstracted at various levels and allows experimenting with both single (Bot) and multiple collaborative agents (CoBot). The class diagram for the same is given in Fig. \ref{fig:class_diagram}. It was kept in mind that error between RobotSim\cite{vasarhelyi2018optimized} and the current work should remain under a threshold of 5-10\%. The link of the repository for the code is given in supplementary material (S1) and a screenshot of the simulation is shown in Fig. \ref{sim_screenshot}.
All the experiments are carried out with a flocking velocity ($v^{flock}$) and maximum velocity ($v^{max}$) of 6 m/s but no changes were made in the algorithm so as to disrupt the scalability in velocity. Artificial GPS noise is added using the Brownian noise model used in \cite{viragh2014flocking}. Communication delays are integral to the result of optimization as they simulate a kind of inertia at the walls and with neighbours as well. Without these delays and noises it is observed that the drones favour high gain and short range repulsion as opposed to the model optima. 

After analysing the covariance matrix for correlations (Section \ref{pca}), the objectives are combined accordingly and passed to the multi-objective optimizer.
A good multi-objective optimization algorithm should contain the following characteristics:

\begin{enumerate}
    \item Guide the solutions to an optimal Pareto front
    \item Maintain solution diversity
\end{enumerate}

NSGA-II is proven to be one of the best performing algorithms in this regard. The ‘pymoo’~\cite{blank2020pymoo} python library is used for the same with default parameters.

It is imperative to setup the optimization problem in such a way that there is enough diversity in the search space so as to find "good enough" solutions through heuristic methods. For the sake of exploration a test run is conducted using the CMA-ES algorithm without any bounds on the parameters. The solution for this setup revealed that the flock only moves in circles around the centre  and does not interact with the walls at all. While such a solution is mathematically the most optimal, it does not encapsulate the physical limitations and logical constraints on the variable bounds. This happens because the correlation and wall fitnesses become abnormally high.  

To avoid such false positives in the simulation, either explicitly known bounds can be set on the variables which are realistic and relevant to the physics of a UAV or another objective which maximises the search area covered in minimum time can also be incorporated in the optimization process. For this study, the former approach is used without any loss of generality. The bounds used for the variables are shown in Table \ref{tab_bounds} and some miscellaneous simulation parameters including certain tolerance parameters $r^{tol}$, $a^{tol}$, and $v^{tol}$ for the transfer functions in section \ref{sec:fitfuncs} are given in Table \ref{tab_params}. Appropriate values for these tolerance parameters promote better search of solutions and gradient directions.

All the experiments were performed on a machine with the AMD Ryzen 7 4800H 16 core CPU and 16 GB of RAM. The results are reported in Section \ref{results}.

\begin{table}
    \parbox{.49\linewidth}{
    \centering
    \caption{Optimization bounds \label{tab_bounds}}
    \begin{tabular}{lcc} 
    \toprule
    & Lower bound  & Upper bound \\
    \midrule \\
    $r_{0}^{rep}$ & 30.8 & 51 \\
    $p^{rep}$ & 0.02 & 0.10 \\
    $r_{0}^{frict}$ & 58.5 & 100   \\ 
    $ a^{frict}$  & 5.04 & 10.0  \\
    $p^{frict}$ &  0.38 & 9.67   \\
    $v^{frict}$ &  0.3 & 2.7    \\
    $c^{frict}$ &  0.03 & 0.22    \\
    $r_{0}^{shill}$ & -10 & 0  \\
    $v^{shill}$ &  10.0 & 15.0    \\
    $a^{shill}$ &  1.54& 6.55   \\
    $p^{shill}$ &  0.48 & 9.96   \\
    $c^{shill}$ &  0.3 & 1\\
    \botrule
    \end{tabular}
    }
    \hfill
    \parbox{.49\linewidth}{
    \centering
    \caption{Simulation parameters \label{tab_params}}
    \begin{tabular}{lcc} 
    \toprule
    Parameter & Value \\
    \midrule \\
    $v^{flock}$ & 6 m/s \\
    $v^{max}$ & 6 m/s \\
    $t_{del}$ & 0.2 s  \\ 
    $N$ & 30  \\
    $L^{arena}$ & 500 m  \\ 
    $\sigma_{inner}$  & 0.005 $m^{2}/s^{2}$ \\
    $t_{del}$ &  1 s  \\
    $r^{coll}$ &  3 m   \\
    $v^{tol}$ &  3.75 m/s  \\
    $a^{tol}$ & 0.0003\\
    $r^{tol}$ &  5 m     \\
    \botrule 
    \end{tabular}
    }
\end{table}

\section{Results and Discussions} \label{results}
The results of the optimization procedure are analyzed and discussed in this section.
\begin{figure}
    \centering
    \includegraphics[scale=0.4]{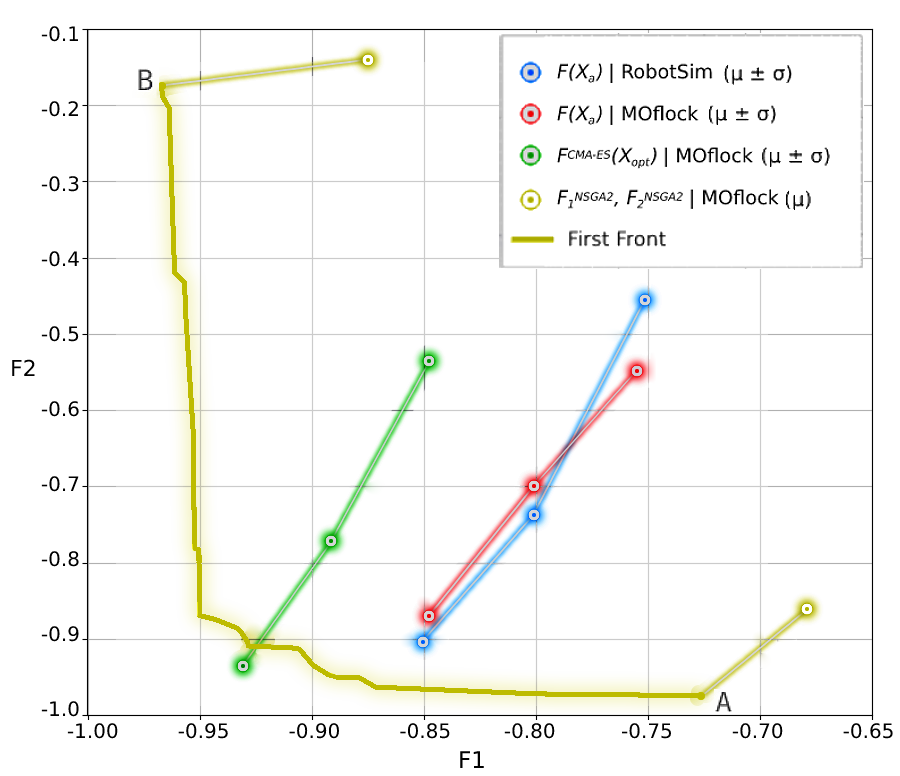}
    \caption{Comparison of different configurations with respect to the optimal Pareto front.}
    \label{fig:optimal_front}
\end{figure}

Fig. \ref{fig:optimal_front} shows statistical evaluations (mean $\pm$ standard deviation) of 100 simulations for different points. The blue and red curves are comparisons of our simulator and RobotSim at the model optima for $v^{flock} = 6 m/s$. $F(X_{a}) \vert \emph{RobotSim}$  is the multi-objective fitness for the model optima $X_{a}$ taken from \cite{vasarhelyi2018optimized} and evaluated on RobotSim itself. $F(X_{a}) \vert \emph{MOflock} $ is the fitness for $X_{a}$ evaluated on MOflock. As targeted, the error on mean fitnesses between both simulator at the model optima in \cite{vasarhelyi2018optimized} is 4.28\%. $F^{CMA-ES}(X_{opt}) \vert \emph{MOflock}$ is the optimized fitness result on out simulator using the CMA-ES algorithm. Note that this fitness wasn't evaluated using a multi-objective algorithm but was separated into $F_{1}$ and $F_{2}$ according to Section \ref{pca}. This is done so that comparisons can be drawn easily between the single objective and multi-objective results. The Pareto front for the last generation using the NSGA-II algorithm is also shown. $F_{1}^{NSGA2}, F_{2}^{NSGA2} \vert \emph{MOflock} $ are mean values for the extreme points on this Pareto Front.

\begin{figure}
    \centering
    \begin{subfigure}[b]{0.53\textwidth}
    \centering
        \includegraphics[width=\textwidth]{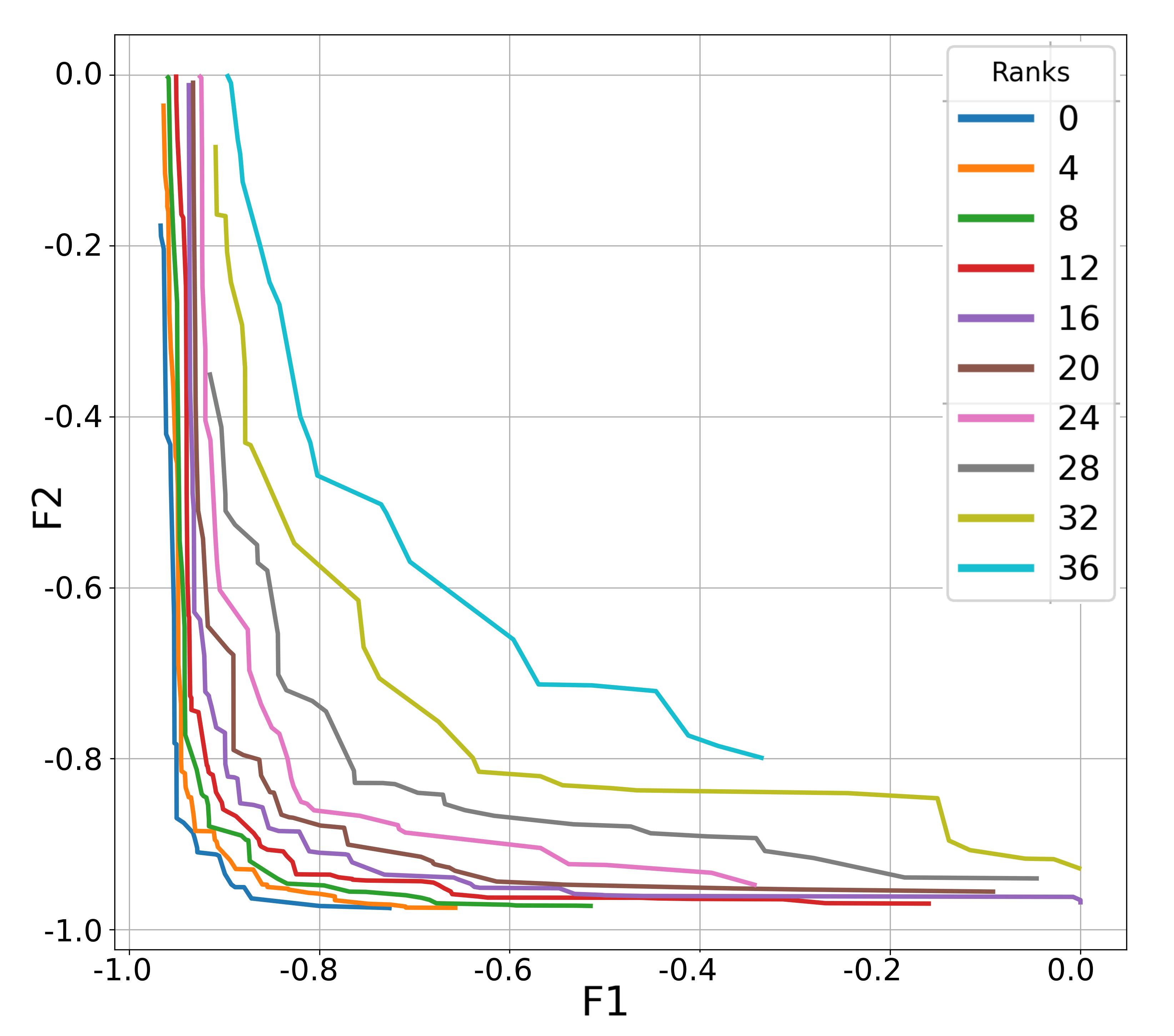}
        \caption{Last 10 even Pareto fronts}
        \label{fig:last_fronts}
    \end{subfigure}
    % \hfill
    \begin{subfigure}[b]{0.46\textwidth}
        \centering
        \includegraphics[width=\textwidth]{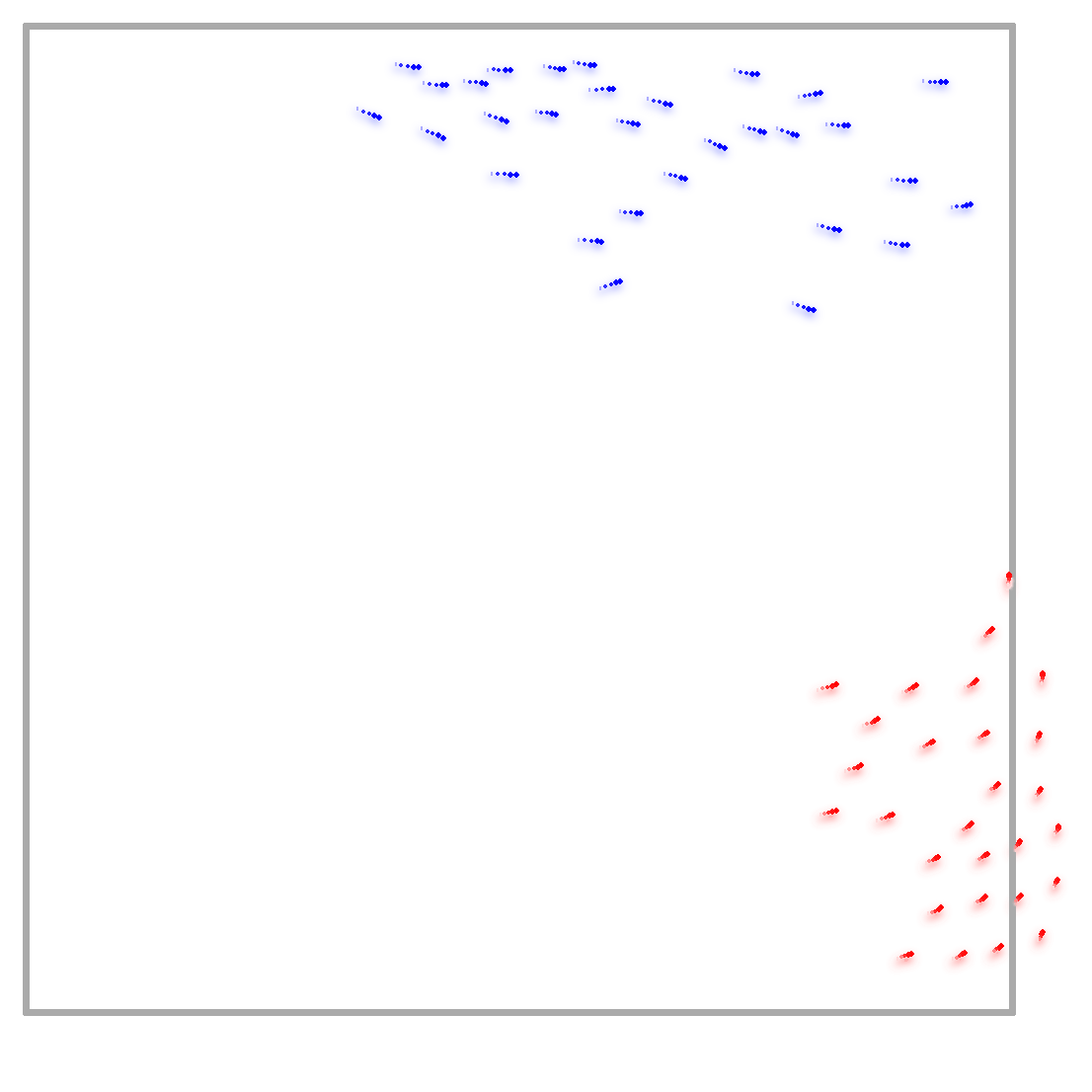}
        \caption{Point A (blue) and Point B (red)}
        \label{fig:sim_a_b}
    \end{subfigure}
    \centering
    \caption{Optimization Results}
\end{figure}

Since the single objective fitness is just a scalar product of all individual fitnesses, it follows that neither of the six fitnesses can be close to zero or even guaranteed to be maximum in case there exists a negative correlation between some of them. As a result, when optimizing a single objective function a 'best of both' situation is sought after. In the case of multiple conflicting objectives, however, this can be forgiven for better performance on the separated fitness functions. This also explains why the CMA-ES point lies around the knee of the Pareto fronts. It should be noted however, that the CMA-ES optima on our simulator outperforms the Pareto front at it's knee. This is owed to the high degree of automation and robustness of the CMA-ES algorithm.

\begin{figure}
    \centering
    \begin{subfigure}[b]{0.49\textwidth}
    \centering
        \includegraphics[width=\textwidth]{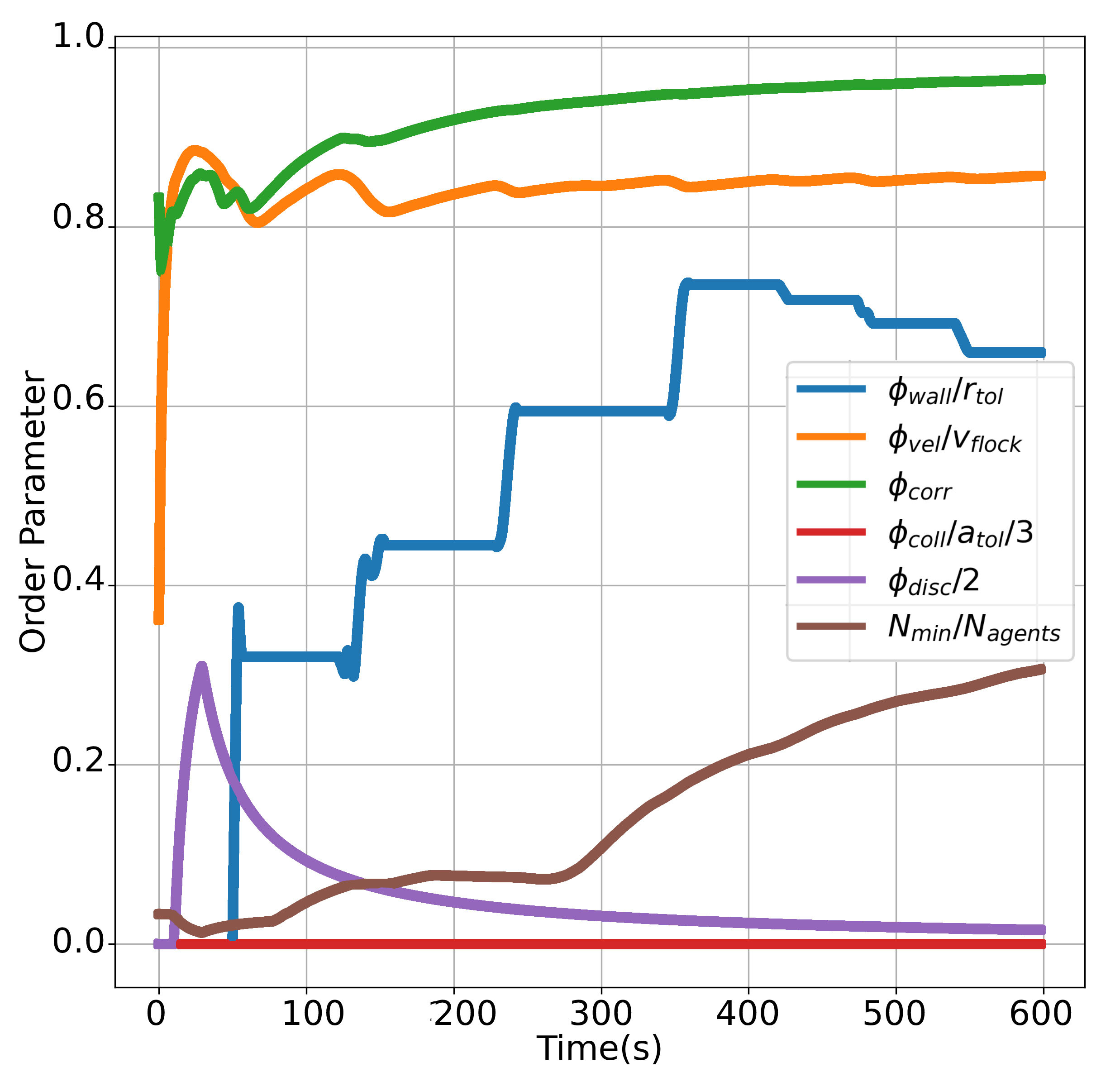}
        \caption{Point A}
        \label{fig:op_a}
    \end{subfigure}
    % \hfill
    \begin{subfigure}[b]{0.49\textwidth}
        \centering
        \includegraphics[width=\textwidth]{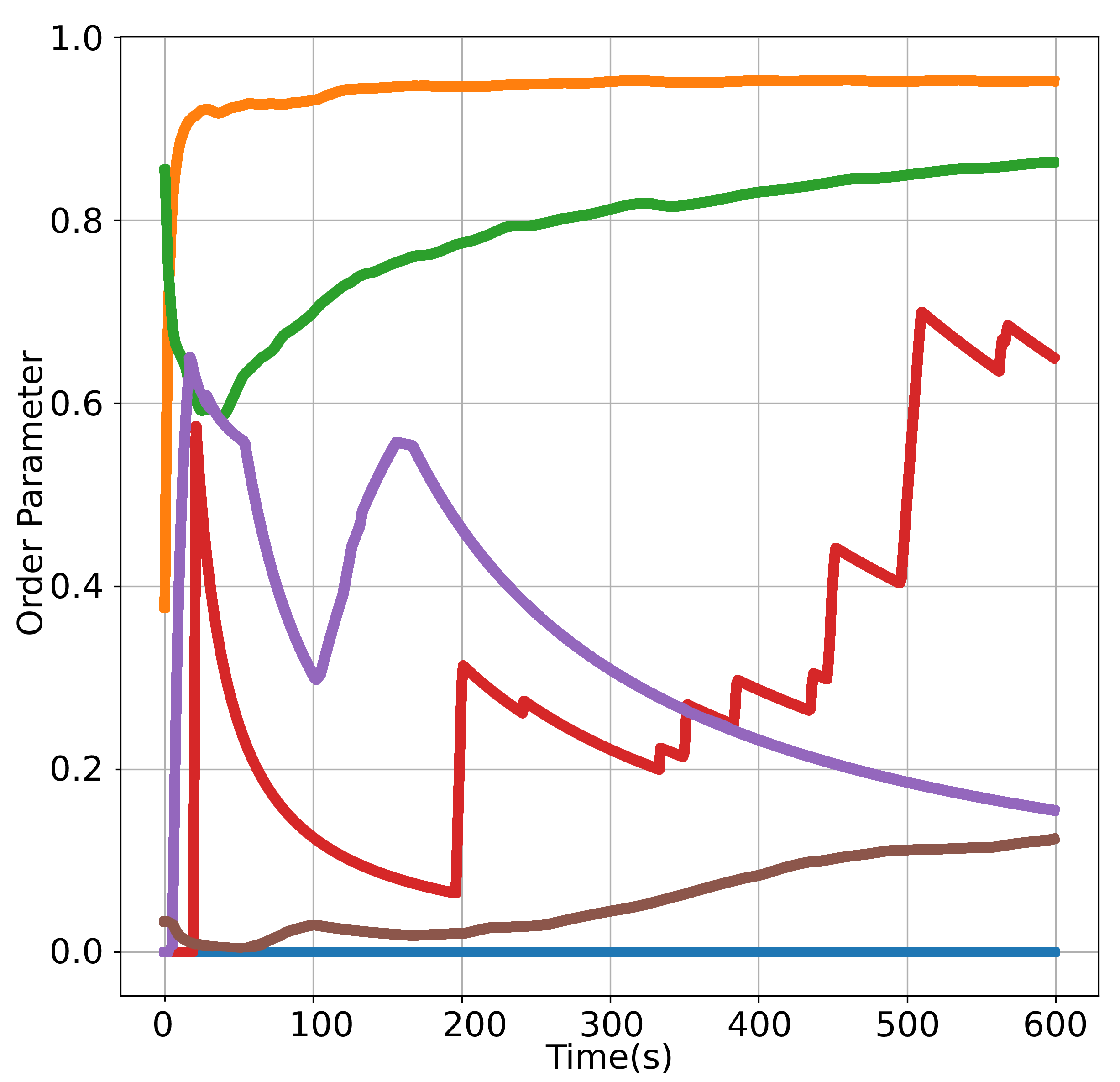}
        \caption{Point B}
        \label{fig:op_b}
    \end{subfigure}
    \centering
    \caption{Cumulative order parameters for points A and b.}
    \label{fig:op_comparison}
\end{figure}

While the user can now choose amongst any of the points depending on the scenario and relative importance, there are two interesting points on the optimal front corresponding to the extreme situations when either one of the two solutions is compromised for the other. They are given by point A and B in Fig. \ref{fig:optimal_front}. The values of the variables and fitnesses at the above points are summarised in Table \ref{tab_results}. Fig. \ref{fig:last_fronts} shows the last 10 even Pareto fronts ranked in descending order. The even ones were only chosen to display the spread and convergence in a neat manner. A snapshot of the relevant simulations for both points is also shown in Fig. \ref{fig:sim_a_b} along with the graphs for their order parameters in Fig. \ref{fig:op_comparison}. They can be qualitatively understood as follows:

Point A: A weaker cluster dependent fitness shows that multiple clusters can coexist in the same environment when correlation and speed is sacrificed.

Point B: Similarly, the other point clearly skips the geo-fence and/or slows down to maintain a good correlation and make up for the damping caused by inter-agent friction and pressure at the walls.

The generation of the above two points is a direct consequence of the physical and environmental restrictions imposed on the swarm. The limited acceleration does not allow the entire swarm to turn sharply without slowing down. The confined walls don’t allow agents to flock together when moving at high speeds without losing on some amount of correlation. These statements are a testament to the complex dynamics that multi-agent systems exhibit. A video showing the above interactions is given in supplementary material (S2). Better mathematical formalism and high fidelity simulations can be developed to realise such intertwined relationships.

The trend in the order parameters  in Fig. \ref{fig:op_comparison} also confirms the elements of the covariance matrix in Section \ref{pca}. Note that the graph is scaled to the (0,1) interval with the relevant maximum feasible values for each parameter and cumulative values are shown for the curves.

\begin{table}[h]
    \centering
    \caption{Optimization results}\label{tab_results}
    \begin{tabular}{l c c}
        \toprule
        & Point A  & Point B\\
        \midrule 
        $F_{1}$($\mu \pm \sigma$) & $0.890 \pm 0.039$ & $0.065 \pm 0.039$ \\
        $F_{2}$($\mu \pm \sigma$) & $0.112 \pm 0.09$ & $0.896 \pm 0.179$ \\
        \midrule
        $r_{0}^{rep}$ & 33.69 &    33.45     \\
        $p^{rep}$ & 0.023   & 0.028    \\
        $r_{0}^{frict}$ & 59.26    & 58.95    \\ 
        $ a^{frict}$  & 5.38   & 8.223  \\
        $p^{frict}$ & 4.62  & 2.67    \\
        $v^{frict}$ & 1.73  & 3.00    \\
        $c^{frict}$ & 0.035  & 1.84    \\
        $r_{0}^{shill}$ & -2.45   & -0.21  \\
        $v^{shill}$ & 12.93 &  12.93    \\
        $a^{shill}$ & 4.84  & 2.57   \\
        $p^{shill}$ & 4.83  & 1.30   \\
        $c^{shill}$ & 0.55 & 0.43 \\
        \botrule
    \end{tabular}
\end{table}

Further statistical analysis on the data from the optimization shows that there is a lot of redundancy in the decision variables. The following observations indicate this finding:
\begin{itemize}
    \item Even though points A and B are far apart on the Pareto front, their respective parameters for repulsion are very similar. \item It was observed that the right combination of $r_{0}^{shill}$ and $a^{shill}$ gives similar fitnesses and order parameters even with a constant shilling velocity. 
    \item The introduced shill gain ($c^{shill}$) doesn't take it's maximum possible value (1.0) even when seeking the best $F_{1}$ which is highly dependent on this parameter.
\end{itemize}
Note that a full PCA correlation analysis on the decision variables can be performed to confirm the above observation, and reduce the dimension of the input space as well.

The above results are more consequential than just a Pareto front. Real life missions and the inherent stochastic nature of the environment demands a range of potential solutions from which a human in the loop can choose in an ad-hoc manner. A typical mission profile consists of cruise, loiter, surveillance, and occasionally a payload drop. A brief description of the use of the practical applications of the Pareto optimal points are shown below.

\begin{figure}
    \centering
    \begin{subfigure}[b]{0.48\textwidth}
    \centering
        \includegraphics[width=\textwidth]{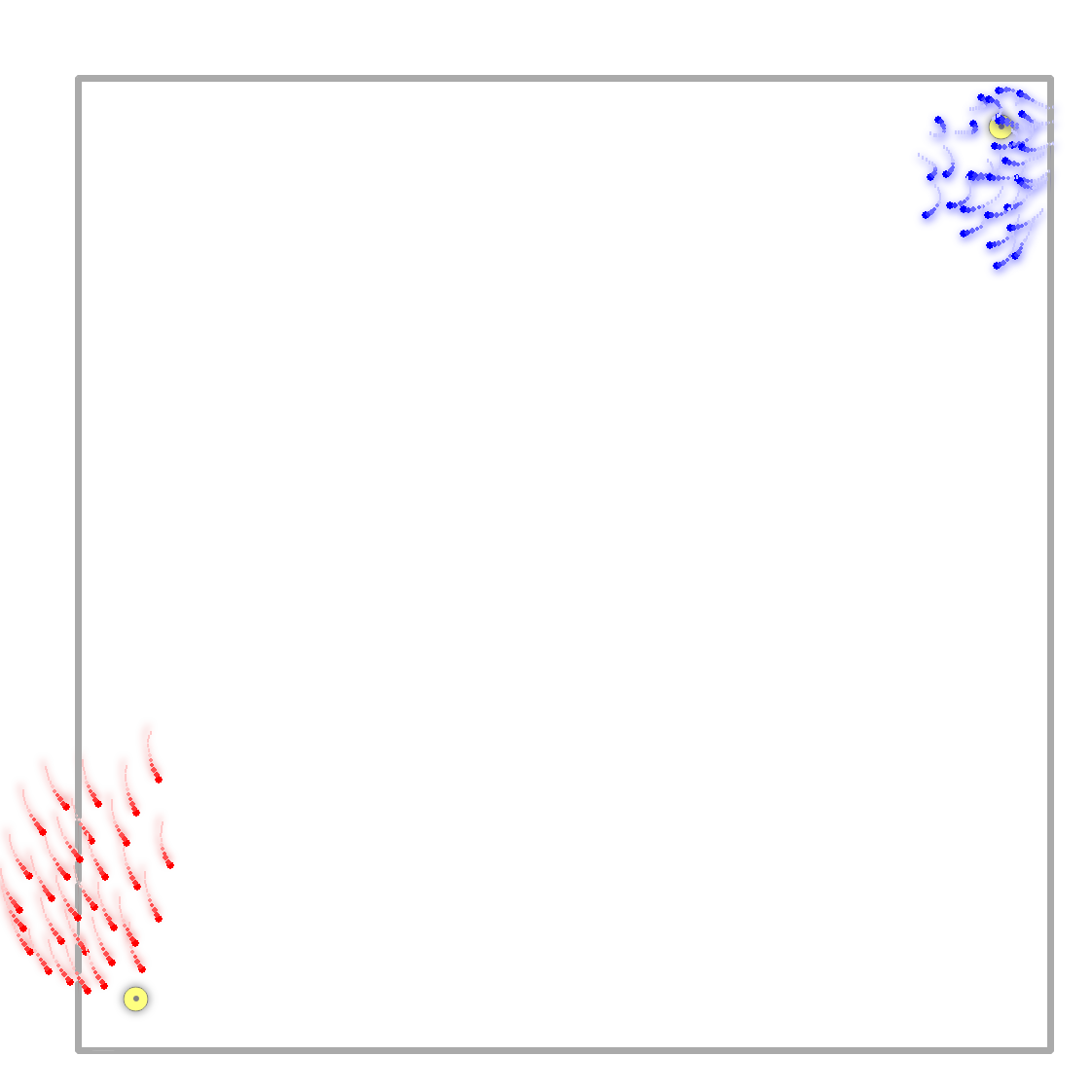}
        \caption{Target following simulation}
        \label{fig:target_sim}
    \end{subfigure}
    % \hfill
    \begin{subfigure}[b]{0.51\textwidth}
        \centering
        \includegraphics[width=\textwidth]{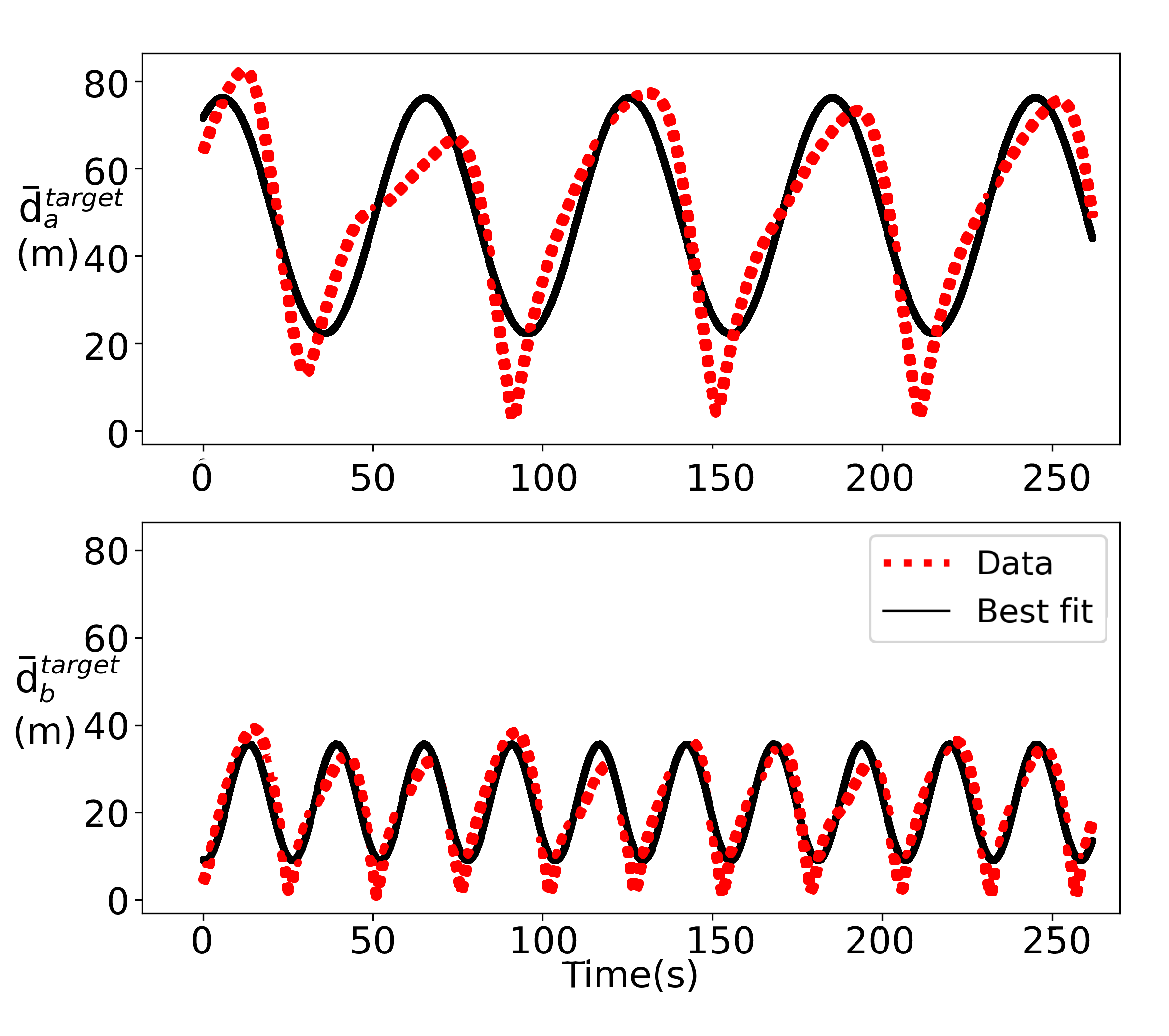}
        \caption{$\bar{d}^{target}(t)$ for Point A and B}
        \label{fig:target_curve}
    \end{subfigure}
    \centering
    \caption{Target order parameter comparison}
    \label{fig:target_op_comparison}
\end{figure}

\begin{itemize}
    \item Target search and loitering is a common phase in surveillance missions. A snapshot of an extreme case where the target is located at a corner of the geo-fence is shown in Fig. \ref{fig:target_sim}. The flock breaks at corners and walls to loiter around the target. To make this observation mathematically sound, another order parameter called $\phi^{target}$ is created.
\end{itemize}
\begin{align}
    \textbf{x}^{COM}(t) &= \frac{\sum_{i =1}^{N} \textbf{r}_{i}(t)}{N}\\
    \bar{d}^{target}(t) &= \| \textbf{x}^{target}-\textbf{x}^{COM}(t) \| \label{eq_dtarget}\\
\end{align}

\begin{flushleft}
    where,\\
    $\textbf{x}^{COM}(t) \equiv$ Center of mass of the swarm at time t\\
    $\bar{d}^{target}(t) \equiv$ Mean distance to target (over all $N$ agents) at time t \\ 
\end{flushleft}

This parameter includes two performance measures- the closeness of the entire flock to the target on average (Eq. \ref{eq_dtarget}) and the ‘Loiter Frequency ($\omega$)’. This frequency measures how fast the flock can loiter around the target and turn around as a whole. As opposed to the other parameters, the steady state version of this parameter is measured. Since the motion is circular and periodic, the time series is fit to a sinusoidal wave similar to an audio signal.

\begin{align}
    \phi^{target}(t) &= a. \sin(\omega. \bar{d}^{target}(t) + \psi ) + c  \label{eq_phitarg}\\
    F &= FFT(\phi^{target})  \label{eq_fft} \\
    \overline{\bar{d}^{target}} &= \frac{1}{T} \sum_{t=0}^{T-1} \bar{d}^{target}(t)\\
    a_{o} &= \sqrt{\frac{2}{T}\sum_{t=0}^{T-1}(\bar{d}^{target}(t)- \overline{\bar{d}^{target}})^{2}}\\
    f_{o} &= \vert f^{s}_{argmax( \vert A_{k} \vert ) } \vert \\
    \psi_{o} &= 0\\
    c_{o} &=  \overline{\bar{d}^{target}}\\
    a, \omega, \psi, c &= LSF(\phi^{target}, \bar{d}^{target},a_{o}, f_{o}, \psi_{o}, c_{o}) \label{eq_lsf}
\end{align}

\begin{flushleft}
    where,\\
    $F \ \equiv $ Fourier transform output\\
    $\overline{\bar{d}^{target}} \equiv$ Mean of the mean distance throughout the simulation \\ 
    $f^{s} \ \equiv$ Sample frequencies for the time series data\\
    $\bar{d}^{target} \ \equiv $  Average distance throughout simulation \\
    $f_{o} \ \equiv $  Initial guess of frequency for $\phi^{target}(t)$ corresponding to the maximum $F$ \\
    $\psi_{o} \ \equiv$ Initial guess of phase for $\phi^{target}(t)$\\
    $a_{o} \ \equiv$ Initial guess of amplitude for $\phi^{target}(t)$\\
    $c_{o} \ \equiv$ Initial guess of offset for $\phi^{target}(t)$
\end{flushleft}

\sloppy
This is done by first getting an estimate of the initial coefficients, namely amplitude ($a_{o}$), phase ($\psi_{o}$), offset ($c_{o}$), and frequency ($f_{o}= \omega_{o}/2\pi)$) via a Fast Fourier Transform ($FFT$) on the data (Eq. \eqref{eq_phitarg}-\eqref{eq_fft}) and then passing this estimate for Least Squares curve Fit represented by $LSF$ (Eq. \eqref{eq_lsf}). The final order parameter is just the angular frequency divided by the amplitude. 
\begin{align}
    F^{target} &= \omega/a
\end{align}

The analysis shows that point A on the Pareto front has a lower loiter frequency because of the extra inter-agent friction created to maintain the flock correlation. Point B on the other hand has almost half the amplitude and double the frequency because of the higher velocity, loosely correlated flock with more collisions. These curves and an accompanying simulation screenshot are shown in Fig. \ref{fig:target_op_comparison}. A full video showing the target tracking and fitness analysis is given in supplementary material S3.

\begin{itemize}
    \item There have been recent studies in which collisions are handled explicitly through physical boundaries and mechanisms rather than by an explicit algorithm \cite{mulgaonkar2017robust}. The idea is to allow for some amount of collisions as long as agility is maintained and the drones reach their target. Point A on the front is akin to such a situation. The flock doesn't give much attention to inter-agent separation or correlation in local clusters. Rather, the speed is  given a higher priority. This is especially useful when tiny drones need to overcome narrow passages and crevices without acting as a fully connected flock but get through the region as fast as possible with each drone acting for themselves.
    \item Point B naturally resembles a good flock where connectivity and correlation is concerned. The decentralised neighbor architecture makes the flock very desirable where robustness and swarm health is an absolute requirement and the entire swarm needs to travel long distances as a fully connected cluster.
\end{itemize}

While developing the methodology for this work, there were a number of nuanced characteristics of collective behaviour noticed in the multi-agent simulations. For instance, the parameters which characterise the swarm changed drastically based on factors like communication delay and the arena size. These two variables affect the swarm as a whole because any control action for an agent close to the wall is propagated throughout the swarm with the appropriate communication delay. Naturally, every PCA analysis with different simulation parameters yielded unique objectives and therefore a different Pareto front. The advantage of separating the objective function into multiple grouped objectives is that global swarm behaviour can be controlled  by choosing a point on the Pareto front instead of tuning parameters manually or running an offline optimization for each possible situation that the swarm would encounter. It follows therefore, that the behaviour of the swarm can be controlled by a supervisor with access to the appropriate Pareto front. This problem of generalising a semi-autonomous swarm based on various scenarios is often difficult to handle with just online learning algorithms. A compromise between both, wherein, we can control large scale behaviour through multiple objectives, and individual decision making through reinforcement learning can be sought after to solve the generalisation problem.

\section{Conclusion}

In this paper, we proposed a methodology to address the problem of drone flocking. First, a simulator with an integrated optimizer was designed to test the algorithm. The decision variables which characterise the flocking operators, and fitness functions which indicate the performance of the swarm are defined. Then, to use the multi-objective optimizer effectively, the six dimensional objective space is reduced to two dimensions using Principal Component Analysis. The correlation analysis showed that fitness functions for both speed and wall avoidance can be treated separately from the cohesive movement of the entire flock. This process also gave insight into the various complex relationships that multi-agent systems can exhibit. Further, the so formed two objective optimization problem is optimized using NSGA-II and the results are compared with the single objective CMA-ES optimization algorithm. It is found that while CMA-ES performs better with respect to the knee of the Pareto front, NSGA-II outperforms CMA-ES on the extreme points and offers an entire range of solutions to choose from. The study also discussed the use cases of such a Pareto front to guide the decision-making process in real-world scenarios. Incorporating algorithms like Reinforcement Learning with the proposed methodology can be future research agenda.

\section*{Supplementary Material}
\begin{itemize}
    \item S1: Repository: Multi-objective flocking simulator (\url{https://github.com/nikhil-sethi/MOflock})
    \item S2: Comparison of the swarms from the two extreme points on the Pareto front (\url{https://youtu.be/MIgc80M1dHg}
    \item S3: Target following for the two extreme points on the Pareto front (\url{https://youtu.be/Fl71fg-DU_c})
\end{itemize}

\section*{Data Availability}
The datasets generated during and/or analysed during the current study are available from the corresponding author on reasonable request.

\section*{Declarations}
This work was supported by the funding received in the international joint research project titled “Design and Applications of Swarm Intelligence based algorithms for drone swarm and COVID19 Spread Prediction” funded by Liverpool Hope University UK. The authors declare that there is no conflict of interest.

\bibliography{ref}
\end{document}